**Title:**

# Lexical Knowledge Representation in an Intelligent Dictionary Help System

**Authors and affiliation:**

**E. Agirre, X. Arregi, X. Artola, A. Díaz de Ilarraza, K. Sarasola**

Informatika Fakultatea (Univ. of the Basque Country)
P. K. 649, 20080 DONOSTIA (Basque Country - Spain)
E-mail: jiparzux@si.ehu.es

## Summary

The frame-based knowledge representation model adopted in IDHS (Intelligent Dictionary Help System) is described in this paper. It is used to represent the lexical knowledge acquired automatically from a conventional dictionary. Moreover, the enrichment processes that have been performed on the Dictionary Knowledge Base and the dynamic exploitation of this knowledge —both based on the exploitation of the properties of lexical semantic relations— are also described.

cmp-lg/9501004    30 Jan 1995

# LEXICAL KNOWLEDGE REPRESENTATION IN AN INTELLIGENT DICTIONARY HELP SYSTEM

E. Agirre, X. Arregi, X. Artola, A. Díaz de Ilarraza, K. Sarasola

Informatika Fakultatea (Univ. of the Basque Country)
P. K. 649, 20080 DONOSTIA (Basque Country - Spain)    E-mail: jiparzux@si.ehu.es

## 1 INTRODUCTION.

IDHS (Intelligent Dictionary Help System) is conceived as a monolingual (explanatory) dictionary system for human use (Artola & Evrard, 92). The fact that it is intended for people instead of automatic processing distinguishes it from other systems dealing with semantic knowledge acquisition from conventional dictionaries. The system provides various access possibilities to the data, allowing to deduce implicit knowledge from the explicit dictionary information. IDHS deals with reasoning mechanisms analogous to those used by humans when they consult a dictionary. User level functionality of the system has been defined and is partially implemented.

The starting point of IDHS is a Dictionary Database (DDB) built from an ordinary French dictionary. Meaning definitions have been analysed using linguistic information from the DDB itself and interpreted to be structured as a Dictionary Knowledge Base (DKB). As a result of the parsing, different lexical-semantic relations between word senses are established by means of semantic rules (attached to the patterns); this rules are used for the initial construction of the DKB.

This paper describes the knowledge representation model adopted in IDHS to represent the lexical knowledge acquired from the source dictionary. Once the acquisition process has been performed and the DKB built, some enrichment processes have been executed on the DKB in order to enhance its knowledge about the words in the language. Besides, the dynamic exploitation of this knowledge is made possible by means of specially conceived deduction mechanisms. Both the enrichment processes and the dynamic deduction mechanisms are based on the exploitation of the properties of the lexical semantic relations represented in the DKB.

In the following section an overview of IDHS is given. Section 3 briefly presents the process of construction of the DKB. The knowledge representation model and the enrichment mechanisms are fully described in sections 4 and 5. Section 6 describes some inferential aspects of the system. Finally, in section 7, some figures about the size of the prototype built are presented.

## 2 THE IDHS DICTIONARY SYSTEM.

IDHS is a dictionary help system intended to assist a human user in language comprehension or production tasks. The system provides a set of functions that have been inspired by the different reasoning processes a human user performs when consulting a conventional dictionary, such as definition queries, search of alternative definitions, differences, relations and analogies between concepts, thesaurus-like word search, verification of concept properties and interconceptual relationships, etc. (Arregi *et al.*, 91).

IDHS can be seen as a repository of dictionary knowledge apt to be accessed and exploited in several ways. The system has been implemented on a symbolic architecture machine using KEE knowledge engineering environment.

Two phases are distinguished in the construction of the DKB. Firstly, information contained in the DDB is used to produce an initial DKB. General information about the entries obtained from the DDB (POS, usage, examples, etc.) is conventionally represented —attribute-value pairs in the frame structure— while the semantic component of the dictionary, i.e. the definition sentences, has been analysed and represented as an interrelated set of concepts. In this stage the relations established between concepts could still be, in some cases, of lexical-syntactic nature. In a second phase, the semantic knowledge acquisition process is completed using for that the relations established in the initial DKB. The purpose of this phase is to perform lexical and syntactical disambiguation, showing that semantic knowledge about hierarchical relations between concepts can be determinant for this.

## 3 BUILDING THE DICTIONARY KNOWLEDGE BASE.

The starting point of this system is a small monolingual French dictionary (*Le Plus Petit Larousse*, Paris: Librairie Larousse, 1980) consisting of nearly 23,000 senses related to almost 16,000 entries. The dictionary was recorded in a relational database: the Dictionary Database (DDB). This DDB is the basis of every empirical study that has been developed in order to design the final model proposed for representation and intelligent exploitation of the dictionary.

The definition sentences have been analysed in the process of transformation of the data contained in the DDB to produce the DKB. The analysis mechanism used is based on hierarchies of phrasal patterns (Alshawi, 89). The semantic structure associated to each analysis pattern is expressed by means of a Semantic Structure Construction Rule (SSCR). The process of construction of the DKB is automatic and based on these SSCR's (Artola, 93).

The interconceptual lexical-semantic relations detected from the analysis of the source dictionary are classified into paradigmatic and syntagmatic. Among the paradigmatic relations, the following have been found: synonymy and antonymy, taxonomic relations as hypernymy/hyponymy —obtained from definitions of type "genus et differentia"—, and taxonymy itself (expressed by means of specific relators such as *sorte de* and *espèce de*), meronymy, and others as gradation (for adjectives and verbs), equivalence (between adjectives and past participle), factitive and reflexive (for verbs), lack and reference (to the previous sense). Whereas among the syntagmatic relations, i.e. those that relate concepts belonging to different POS's, derivation is the most important, but also relationships between concepts without any morphological relation as case relations, attributive (for verbs), lack and conformity have been detected.

The hierarchies created have already been used to parse all the noun, verb, and adjective definitions in the DDB. The hierarchy devoted to analyze noun definitions is formed with 65 patterns, 49 different patterns have been defined to analyze verb definitions, and 45 for adjectives. Although it is a partial parsing procedure, 57.76% of noun definitions, 79.8% of verbs and 69.04% of those corresponding to adjectives have been totally "caught" in this application. However, with this technique of partial parsing, the parse is considered successful when an initial phrase structure is recognized, which in general contains the genus or superordinate of the defined sense. This is not so for the case of lexicographic meta-language constructions (specific relators), whose corresponding semantic structure is built in a specific way and which deserve also specific patterns in the hierarchies.

## 4 REPRESENTATION OF THE DICTIONARY KNOWLEDGE: THE DKB.

As we have just seen, the knowledge representation scheme chosen for the DKB of IDHS is composed of three elements, each of them structured as a different knowledge base:

- KB-THESAURUS is the representation of the dictionary as a semantic network of frames, where each frame represents a *one-word concept* (word sense) or a *phrasal concept*. Phrasal concepts represent phrase structures associated to the occurrence of concepts in meaning definitions. Frames —or units— are interrelated by slots representing lexical-semantic relations such as synonymy, taxonomic relations (hypernymy, hyponymy, and taxonymy itself), meronymic relations (part-of, element-of, set-of, member-of), specific relations realised by means of meta-linguistic relators, casuals, etc. Other slots contain phrasal, meta-linguistic, and general information.
- KB-DICTIONARY allows access from the dictionary word level to the corresponding concept level in the DKB. Units in this knowledge base represent the entries (words) of the dictionary and are directly linked to their corresponding senses in KB-THESAURUS.
- KB-STRUCTURES contains meta-knowledge about concepts and relations in KB-DICTIONARY and KB-THESAURUS: all the different structures in the DKB are defined here specifying the corresponding slots and describing the slots by means of facets that specify their value ranges, inheritance modes, etc. Units in KB-THESAURUS and KB-DICTIONARY are subclasses or instances of classes defined in KB-STRUCTURES.

Fig. 1 gives a partial view of the three knowledge bases which form the DKB with their correspondent units and their inter/intra relationships.

In the KB-THESAURUS, some of the links representing lexical-semantic relations are created when building the initial version of the knowledge base, while others are deduced later by means of specially conceived deduction mechanisms.

When a dictionary entry like *spatule I 1: sorte de cuiller plate* (a kind of flat spoon) is treated, new concept units are created in KB-THESAURUS (and subsidiarily in KB-DICTIONARY) and linked to others previously included in it. Due to the effect of these links new values for some properties are propagated through the resulting taxonomy.

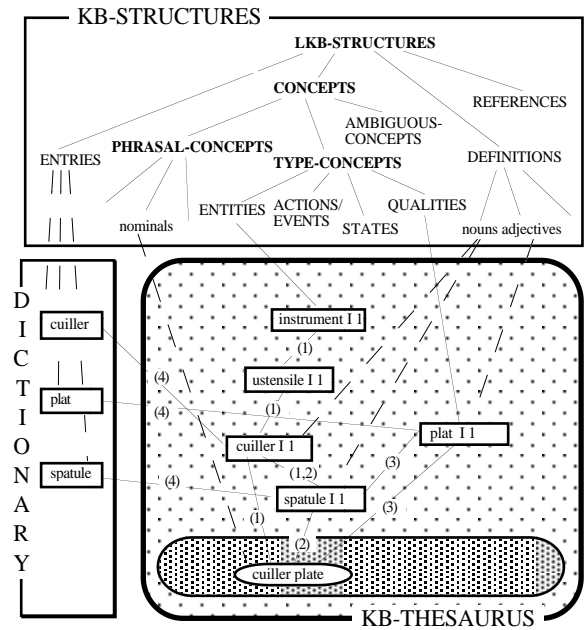

**Fig. 1.-** The Dictionary Knowledge Base.

──── SUBCLASS link
--- MEMBER-OF link (instance)
(1) Taxonomic Relation: HYPERNYM/HYPONYM
(2) Specific (meta-linguistic) relation: SORTE-DE /SORTE-DE+INV (KIND-OF/KIND-OF+INV)
(3) CARACTERISTIQUE /CARACTERISTIQUE+INV (PROPERTY/PROPERTY+INV) relation
(4) MOTS-ENTREE /SENS (ENTRY-WORD / WORD-SENSE) relation

In the example, although it is not explicit in the definition, *spatule* is "a kind of" *ustensile* and so it will inherit some of its characteristics (depending upon the inheritance role of each attribute). Fig. 1 also shows the types of concepts used: *spatule I 1* and *cuiller I 1* are noun definitions and considered subclasses of ENTITIES while *plat I 1* (an adjective) is a subclass of

QUALITIES. The phrasal concept unit representing the noun phrase *cuiller plate* is treated as a hyponym of its nuclear concept (*cuiller I 1*).

### 4.1 KB-STRUCTURES: the meta-knowledge.

This knowledge base reflects the hierarchical organisation of the knowledge included in the DKB.

We will focus on the LKB-STRUCTURES class which defines the data types used in KB-DICTIONARY and KB-THESAURUS, and that organises the units belonging to these knowledge bases into a taxonomy.

Slots defined in KB-STRUCTURES have associated aspects such as the value class, the inheritance role determining how values in children's slots are calculated, and so on. Each lexical-semantic relation —represented by an attribute or slot— has its own inheritance role. For instance, the inheritance role of the CARACTERISTIQUE relation states that every concept inherits the union of the values of the hypernyms for that relation, while the role defined for the SYNONYMES relation inhibits value inheritance from a concept to its hyponyms.

The subclasses defined under LKB-STRUCTURES are the following:

- ENTRIES, that groups dictionary entries belonging to KB-DICTIONARY;
- DEFINITIONS, that groups word senses classified according to their POS;
- REFERENCES, concepts created in KB-THESAURUS due to their occurrence in definitions of other concepts ("definitionless");
- CONCEPTS, that groups, under a conceptual point of view, word senses and other conceptual units of KB-THESAURUS.

The classification of conceptual units under this last class is as follows:

- **TYPE-CONCEPTS** correspond to Quillian's (1968) "type nodes"; this class is, in fact, like a superclass under which every concept of KB-THESAURUS is placed. It is further subdivided in the classes ENTITIES, ACTIONS/EVENTS, QUALITIES and STATES, that classify different types of concepts.
- **PHRASAL-CONCEPTS** is a class that includes concepts similar to Quillian's "tokens" —occurrences of type concepts in the definition sentences—. Phrasal concepts are the representation of phrase structures which are composed by several concepts with semantic content. A phrasal concept is always built as a subclass of the class which represents its head (the noun of a noun phrase, the verb of a verb phrase, and so on), and integrated in the conceptual taxonomy. Phrasal concepts are classified into NOMINALS, VERBALS, ADJECTIVALS, and ADVERBIALS.

For instance, |plante I 1#3| is a phrasal concept (see Fig. 2), subclass of the type concept |plante I 1|, and represents the noun phrase *"une plante d'ornement"*.

- Finally, the concepts that, after the analysis phase, are not yet completely disambiguated (lexical ambiguity), are placed under the class **AMBIGUOUS-CONCEPTS**, which is further subdivided into the subclasses HOMOGRAPHE (e.g. |faculté ? ?|), SENSE (|panser I ?|), and COMPLEX (|donner I 5/6|), in order to distinguish them according to the level of ambiguity they present.

The links between units in KB-THESAURUS and KB-DICTIONARY are implemented by means of slots tagged with the name of the link they represent. These slots are defined in the different classes of KB-STRUCTURES.

The representation model used in the system is made up of two levels:

- *Definitory level*, where the surface representation of the definition of each sense is made. Morphosyntactic features like verb mode, time, determination, etc. are represented by means of facets attached to the attributes. The definitory level is implemented using *representational attributes*. Examples of this kind of attributes are: DEF-SORTED, DEF-QUI, CARACTERISTIQUE and AVEC.
- *Relational level*, that reflects the relational view of the lexicon. It supports the deductive behaviour of the system and is made up by means of *relational attributes*, that may eventually contain deduced knowledge. These attributes, defined in the class TYPE-CONCEPTS, are the implementation of the interconceptual relations: ANTONYMES, AGENT, CARACTERISTIQUE, SORTE-DE, CE-QUI, etc.

### 4.2 KB-DICTIONARY: from words to concepts.

This knowledge base represents the links between each dictionary entry and its senses (see link 4 in Fig. 1).

### 4.3 KB-THESAURUS: the concept network.

KB-THESAURUS stores the concept network that is implemented as a network of frames. Each node in the net is a frame that represents a conceptual unit: one-word concepts and phrasal concepts.

The arcs interconnect the concepts and represent lexical-semantic relations; they are implemented by means of frame slots containing pointers to other concepts. Hypernym and hyponym relations have been made explicit, making up a *concept taxonomy*. These taxonomic relations have been implemented using the environment hierarchical relationship, in order to get inheritance automatically.

Let us show an example. The representation of the following definition

*géranium I 1:* une plante d'ornement

requires the creation of two new conceptual units in THESAURUS: the one which corresponds to the definiendum and the phrasal concept which represents the noun phrase of the definition. Moreover, the units which represent *plante* and *ornement* are to be created also (if they have not been previously created because their occurrence in another definition).

Let us suppose that three new units are created: |géranium I 1|, |plante I 1#3| and |ornement I 1|.

Attributes in the units may contain facets (attributes for the attributes) used in the definitory level to record aspects like determination, genre and so on, but also to establish the relations between definitory attributes with their corresponding relational, or to specify the certainty that the value in a representational attribute has to be "promoted" to a corresponding relational (see below the case of the slot DE in |plante I 1#3|).

Following is given the composition of the frames of these three units at the definitory level of representation (slots are in small capitals whereas facet identifiers are in italics):

|géranium I 1|
  MEMBER.OF: NOMS
  GROUPE-CATEGORIEL: NOM
    *CLASSE-ATTRIBUT: INFO-GENERALE*
  TEXTE-DEFINITION: "une plante d'ornement"
    *CLASSE-ATTRIBUT: INFO-GENERALE*
  DEF-CLASSIQUE: |plante I 1#3|
    *CLASSE-ATTRIBUT: DEFINITOIRES*
    *DETERMINATION: UN*
    *GENRE: F*
    *RELATIONNELS-CORRESPONDANTS: DEFINI-PAR*

|plante I 1#3|
  SUBCLASS.OF: |plante I 1|
  MEMBER.OF: NOMINALES
  TEXTE: "plante d'ornement"
    *CLASSE-ATTRIBUT: INFO-GENERALE*
  DE: |ornement I 1|
    *CLASSE-ATTRIBUT: SYNTAGMATIQUES*
    *RELATIONNELS-CORRESPONDANTS: ORIGINE, POSSESSEUR, MATIERE, OBJECTIF*
    *OBJECTIF: 0.9*

|ornement I 1|
  MEMBER.OF: REFERENCES

Before showing the representation of these units at the relational level, it has to be said that after the initial DKB has been built some deductive procedures have been executed: e.g. deduction of inverse relationships, taxonomy formation, etc. It is to say that in Fig. 2, where the relational view is presented, the relations deduced by these procedures are also represented.

The conceptual units in THESAURUS are placed in two layers (see Fig. 2), recalling the two planes of Quillian. The upper layer corresponds to type concepts whereas in the lower phrasal concepts are placed. Every phrasal concept is placed in the taxonomy directly depending from its nuclear concept, as a hyponym of it.

It is interesting to notice in the figure the relation of *conceptual equivalence* established between |géranium I 1| and |plante I 1#3| (link labelled (3)). These units represent, in fact, the same concept because |plante I 1#3|, standing for *"une plante d'ornement"*, is the definition of |géranium I 1|.

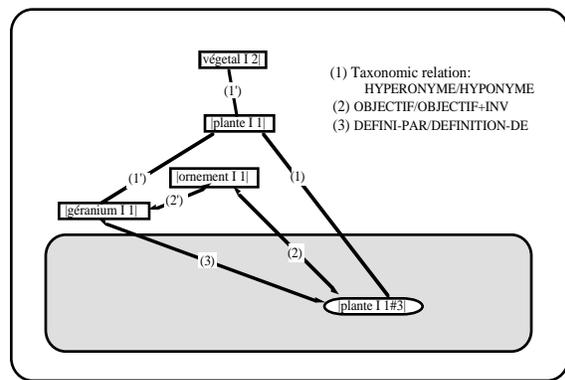

**Fig. 2.** Relational view of the concept |géranium I 1| (in the THESAURUS net).

The frame of |géranium I 1| at the relational level of representation takes the following aspect, once the relational attributes have been (partially) completed:

|géranium I 1|
  SUBCLASS.OF: ENTITES, |plante I 1|
  MEMBER.OF: NOMS
  GROUPE-CATEGORIEL: NOM
    *CLASSE-ATTRIBUT: INFO-GENERALE*
  TEXTE-DEFINITION: "une plante d'ornement"
    *CLASSE-ATTRIBUT: INFO-GENERALE*
  DEF-CLASSIQUE: |plante I 1#3|
    *CLASSE-ATTRIBUT: DEFINITOIRES*
    *DETERMINATION: UN*
    *GENRE: F*
    *RELATIONNELS-CORRESPONDANTS: DEFINI-PAR*
  DEFINI-PAR: |plante I 1#3|
    *CLASSE-ATTRIBUT: RELATIONNELS*
    *INVERSES-CORRESPONDANTS: DEFINITION-DE*
  OBJECTIF: |ornement I 1|
    *CLASSE-ATTRIBUT: RELATIONNELS*
    *INVERSES-CORRESPONDANTS: OBJECTIF+INV*

Let us show now another example. It is the case of two definitions stated by means of two different stereotyped formulae belonging to the lexicographic meta-language. Many verbs in the LPPL are defined by means of a formula beginning with *"rendre"* and many nouns with one beginning with *"qui"*. The definitions selected for this example correspond to the entries ***publier I 1*** and ***ajusteur I 1***, which are represented at the definitory level using the meta-language attributes DEF-RENDRE and DEF-QUI respectively:

***publier I 1:*** *rendre public*
***ajusteur I 1:*** *qui ajuste des pièces de métal*

The frame corresponding to |publier I 1| is the following:

|publier I 1|
  MEMBER.OF: VERBES
  GROUPE-CATEGORIEL: VERBE
    *CLASSE-ATTRIBUT: INFO-GENERALE*
  TEXTE-DEFINITION: "rendre public"
    *CLASSE-ATTRIBUT: INFO-GENERALE*
  DEF-RENDRE: |public I 1|
    *CLASSE-ATTRIBUT: DEFINITOIRES*
    *RELATIONNELS-CORRESPONDANTS: RENDRE*

where it can be seen that no phrasal concept is involved because the link (DEF-RENDRE) is established directly between |publier I 1| and |public I 1|. However, in the case of the definition of *ajusteur I 1*, two phrasal concepts are created: the attribute DEF-QUI points to the phrasal concept |ajuster

I 1#1|, representing *"ajuster des pièces de métal"*, and this phrasal concept, in turn, has a syntagmatic attribute (OBJET) pointing to a nominal that represents *"pièce de métal"*. Let us show the frames involved in this last case:

|ajusteur I 1|
  MEMBER.OF: NOMS
  GROUPE-CATEGORIEL: NOM
    CLASSE-ATTRIBUT: INFO-GENERALE
  TEXTE-DEFINITION: "qui ajuste des pièces de métal"
    CLASSE-ATTRIBUT: INFO-GENERALE
  DEF-QUI: |ajuster I 1#1|
    CLASSE-ATTRIBUT: DEFINITOIRES
    MODE: IND
    ASPECT: NT
    TEMPS: PRES
    PERSONNE: 3
    RELATIONNELS-CORRESPONDANTS: QUI

|ajuster I 1#1|
  SUBCLASS.OF: |ajuster I 1|
  MEMBER.OF: VERBALES
  TEXTE: "ajuster des pièces de métal"
    CLASSE-ATTRIBUT: INFO-GENERALE
  OBJET: |pièce I 1#2|
    CLASSE-ATTRIBUT: SYNTAGMATIQUES
    DETERMINATION: UN
    NOMBRE: PL
    RELATIONNELS-CORRESPONDANTS: THEME

|pièce I 1#2|
  SUBCLASS.OF: |pièce I 1|
  MEMBER.OF: NOMINALES
  TEXTE: "pièce de métal"
    CLASSE-ATTRIBUT: INFO-GENERALE
  DE: |métal I 1|
    CLASSE-ATTRIBUT: SYNTAGMATIQUES
    RELATIONNELS-CORRESPONDANTS: ORIGINE, POSSESSEUR, MATIERE, OBJECTIF
    MATIERE: 0.9

Frequently, phrasal concepts represent "unlabelled" concepts, i.e., they indeed represent concepts that do not have a significant in the language. For instance, there is not, at least in French, a verbal concept meaning *'ajuster des pièces de métal'* nor a noun meaning *'pièce de métal'*. However, this is not the case of the phrasal concepts that are linked to type concepts by means of the relation DEFINI-PAR/DEFINITION-DE, because there, the phrasal concept is, in fact, another representation of the concept being defined (see above the example of the definition of *géranium I 1*). In the representation model proposed in this work, phrasal concepts denote concepts that are typically expressed in a periphrastic way and that do not have necessarily any corresponding entry in the dictionary[1].

Another interesting point related to the creation of these phrasal concepts is the maintenance of direct links between a concept and all the occurrences of this concept in the definition sentences of other concepts. It gives, in fact, a virtual set of usage examples that may be useful for different functions of the final system.

---

[1] This could be very interesting also, in the opinion of the authors, in a multilingual environment: it is possible that, in another language, the concept equivalent to that which has been represented by the phrasal concept |pièce I 1#2| has its own significant, a word that denotes it. In this case, the phrasal concept based representation may be useful to represent the equivalence between both concepts.

## 5 ENRICHMENT PROCESSES PERFORMED ON THE DKB.

In this section the enrichment processes accomplished on the DKB are explained. Two phases are distinguished: (a) the enrichment obtained during the construction of the initial DKB, and (b), where different tasks concerning mainly the exploitation of the properties of synonymy and taxonymy have been performed.

### 5.1 Enrichment obtained during the construction of the initial DKB.

KB-THESAURUS itself, represented —as a network— at the relational level, can be considered an enrichment of the definitory level because, while the DKB was built, the following processes have been performed:

- Values coming from the definitory level have been promoted to the relational level.
- Values coming from the unit which represents the definiens have been transferred to the corresponding definiendum unit.
- The maintenance of the relations in both directions has been automatically guaranteed.
- The concepts included in REFERENCES have been directly related to other concepts.
- The taxonomy of concepts has been made explicit, thus obtaining value inheritance.

### 5.2 Second phase in the enrichment of the DKB.

Several processes have been carried out in order to infer new facts to be asserted in the DKB[2]. The enrichment obtained in this phase concerns the two following aspects:

- Exploitation of the properties of the synonymy (symmetric and transitive).
- Enlargement of the concept taxonomy based on synonymy.

Another aspect that has been considered to be exploited in this phase is that of disambiguation. The use of the lexical-semantic knowledge about hierarchical relations contained in the DKB can be determinant in order to reduce the level of lexical and syntactical ambiguity[3]. Heuristics based on the taxonomic and synonymic knowledge obtained previously have been considered in this phase. Some of them have been designed, implemented and evaluated in a sample of the DKB.

## 6 INFERENTIAL ASPECTS: DYNAMIC DEDUCTION OF KNOWLEDGE.

Dynamic acquisition of knowledge deals with the knowledge not explicitly represented in the DKB and captured by means of especially conceived mechanisms which are activated when the system is to answer a question posed by the user (Arregi *et al.*, 91). The following aspects are considered:

- Inheritance (concept taxonomy).
- Composition of lexical relations.
- Links between concepts and relations: users are allowed to use actual concepts to denote relationships (and not only primitive relations).
- Ambiguity in the DKB: treatment of remaining uncertainty.

---

[2] By means of rules fired following a forward chaining strategy.
3 Lexical ambiguity comes from the definitions themselves; syntactical ambiguity is due mainly to the analysis process.

In the following, some aspects concerning to the second point will be discussed.

In IDHS, the relationships among the different lexical-semantic relations can be easily expressed in a declarative way. It is the way of expressing these relationships that is called the *composition of lexical relations*. From an operative point of view, this mechanism permits the dynamic exploitation —under the user's requests— of the properties of the lexical relations in a direct manner. It is, in fact, a way of acquiring implicit knowledge from the DKB.

The declarative aspect of the mechanism is based on the definition of triples: each triple expresses a relationship among different lexical-semantic relations. These triples have the form ($R_1$ $R_2$ $R_3$), where $R_i$ represents a lexical relation[4]. The operative effect of these declarations is the dynamic creation of transitivity rules based on the triples stated. The general form of these rules is the following:

**if** X $R_1$ Y **and** Y $R_2$ Z **then** X $R_3$ Z

When the value(s) of the attribute $R_3$ are asked, a reading demon (attached to the attribute) creates the rule and fires the reasoning process with a backward-chaining strategy. The deduced facts, if any, will not be asserted in the DKB, but in a temporary context.

For instance, the problem of transitivity in meronymic relations (Cruse, 86; Winston *et al.*, 87) can be easily expressed by stating the triple (PARTIE-DE PARTIE-DE PARTIE-DE) but not stating, for instance, (PARTIE-DE MEMBRE-DE PARTIE-DE), thus expressing that the transitivity in the second case is not true. Examples of other triples that have been stated in the system are:

- Combination of meronymic and non-meronymic relations:

    (PARTIE-DE LOCATIF LOCATIF)
    (LOCATIF HYPERONYME LOCATIF)
    (MEMBRE-DE HYPERONYME MEMBRE-DE)

- Combination of relations derived from the definition meta-language:

    (CARACTERISTIQUE QUI-A POSSESSION)
    (OBJECTIF CE-QUI OBJECTIF)

Explicit rules of lexical composition can be used when the general form of the triples is not valid. These rules are used following the same reasoning strategy.

Following is given the rule derived from the last triple and one instance of it. By means of this rule instance, the fact that the purpose of a *géranium* is the action of *orner* is deduced from the definitions of *géranium* and *ornement*:

   **if**  X OBJECTIF Y **and**  ;;;  the objective of X is Y (entity)
        Y Œ-QUI Z        ;;;  Y "est ce qui" Z (action)
   **then**  X OBJECTIF Z  ;;;  the objective of X is Z (action)

   **if**  |géranium I 1| OBJECTIF |ornement I 1| **and**
        |ornement I 1| Œ-QUI |orner I 1|
   **then**  |géranium I 1| OBJECTIF |orner I 1|

## 7 THE PROTOTYPE OF IDHS: SIZE OF THE DKB.

Following some figures are given in order to show the size of the prototype obtained after the initial construction of the DKB. This prototype contains an important subset of the source dictionary.

KB-DICTIONNAIRE contains 2400 entries, each one representing one word. KB-THESAURUS contains 6130 conceptual units; 1738 units of these are phrasal concepts. In this KB there are 1255 ambiguous concepts. Once the initial construction phase was finished, 19691 relational arcs —interconceptual relationships— had been established.

After the enrichment processes, the number of relational links have been incremented up to 21800 (10.7%). It has been estimated that, using the mechanism of lexical composition, the number interconceptual relations could reach an increment of between 5 and 10%[5].

## 8 CONCLUSIONS.

A frame-based knowledge representation model has been described. This model has been used in an Intelligent Dictionary Help System to represent the lexical knowledge acquired automatically from a conventional dictionary.

The characterisation of the different interconceptual lexical-semantic relations is the basis for the proposed model and it has been established as a result of the analysis process carried out on dictionary definitions.

Several enrichment processes have been performed on the DKB —after the initial construction— in order to add new facts to it; these processes are based on the exploitation of the properties of lexical-semantic relations. Moreover, a mechanism for acquiring —in a dynamic way— knowledge not explicitly represented in the DKB is proposed. This mechanism is based on the composition of lexical relations.

---

[4] The result of the transitivity rule that will be created will be the deduction of values for the $R_3$ attribute. The triples are stored in a facet of $R_3$.

[5] Considering only the set of triples declared until now.